\documentclass[a4paper, 10pt, conference]{ieeeconf}
\makeatletter
\let\NAT@parse\undefined
\makeatother
\usepackage{graphicx}
\usepackage{booktabs}
\usepackage{amsmath}
\usepackage{url}
\usepackage{multicol,lipsum}
\usepackage{mathtools}
\usepackage{amsfonts}
\usepackage{multirow}
\usepackage{bm}
\usepackage{color, colortbl}
\usepackage[colorlinks,bookmarksopen,bookmarksnumbered,citecolor=red,urlcolor=red]{hyperref}
\let\labelindent\relax

\usepackage{enumitem}

\hypersetup
{
	pdftitle = {ContactNet},
	pdfauthor = {Angelo Bratta },
	pdftoolbar = true,
	colorlinks = true,
	linkcolor = black,
	citecolor = black,
	urlcolor = black,
}

\usepackage[usenames,dvipsnames]{xcolor}
\definecolor{blue_iit}{RGB}{51,51,255}
\usepackage{algpseudocode}
\usepackage{algorithm}
\usepackage{glossaries}
\usepackage[tight]{units}
\usepackage[normalem]{ulem} 
\usepackage{cancel}
\definecolor{Gray}{gray}{0.9}
\usepackage{soul}

\newacronym{hyq}{HyQ}{Hydraulically actuated Quadruped}

\newacronym{lf}{LF}{Left-Front}
\newacronym{rf}{RF}{Right-Front}
\newacronym{lh}{LH}{Left-Hind}
\newacronym{rh}{RH}{Right-Hind}

\newacronym{haa}{HAA}{Hip Adduction-Abduction}
\newacronym{hfe}{HFE}{Hip Flexion-Extension}
\newacronym{kfe}{KFE}{Knee Flexion-Extension}

\newacronym{imu}{IMU}{Inertial Measurement Unit}
\newacronym{dofs}{DoFs}{Degrees of Freedom}
\newacronym{rt}{RT}{Real Time}

\newacronym{com}{CoM}{Center of Mass}
\newacronym{cop}{CoP}{Center of Pressure}
\newacronym{zmp}{ZMP}{Zero Moment Point}
\newacronym{icp}{ICP}{Instantaneous Capture Point}
\newacronym{cmp}{CMP}{Centroidal Moment Pivot}
\newacronym{grfs}{GRFs}{Ground Reaction Forces}

\newacronym{ls}{LS}{Least Square}

\newacronym{slip}{SLIP}{Spring Loaded Inverted Pendulum}
\newacronym{eom}{EoM}{Equation of Motions}
\newacronym{qp}{QP}{Quadratic Program}
\newacronym{sqp}{SQP}{Sequential Quadratic Programming}
\newacronym{mic}{MIC}{Mixed-Integer Convex}
\newacronym{cmaes}{CMA-ES}{Covariance Matrix Adaptation Evolution Strategy}
\newacronym{ara}{ARA*}{Anytime Repairing A*}
\newacronym{pca}{PCA}{Principal Component Analysis}
\newacronym{cpg}{CPG}{Central Pattern Generator}
\newacronym{wbc}{WBC}{Whole-Body Control}

\newacronym{pd}{PD}{Proportional-Derivative}

\newacronym{mpc}{MPC}{Model Predictive Control}
\newacronym{nmpc}{NMPC}{Nonlinear Model Predictive Control}
\newacronym{awbc}{c$^3$WBC}{Compliant Contact Consistent Whole-Body Control}
\newacronym{swbc}{sWBC}{Standard Whole-Body Control}
\newacronym{c3wbc}{c$^3$WBC}{Compliant Contact Consistent Whole-Body Control}
\newacronym{ste}{TCE}{Terrain Compliance Estimator}
\newacronym{c3}{\texttt{c}$^3$}{compliant contact consistent}

\newacronym{stance}{STANCE}{\textbf{S}oft \textbf{T}errain \textbf{A}daptation a\textbf{n}d \textbf{C}ompliance \textbf{E}stimation}

\newacronym{wbopt}{WBOpt}{Whole Body Optimization}

\newacronym{hc}{HC}{Hunt and Crossley's}
\newacronym{kv}{KV}{Kelvin-Voigt's}

\newacronym{wllsr}{WLLSR}{Weighted Linear Least Squared Regression}

\newacronym{mae}{MAE}{Mean Absolute Tracking Error}

\newacronym{ode}{ODE}{Open Dynamics Engine}
\newacronym{lip}{LIP}{Linear Inverted Pendulum}
\newacronym{srbd}{SRBD}{Single Rigid Body Dynamics}
\newacronym{mip}{MIP}{Mixed Integer Program}

\newcommand{\Rnum}{\mathbb{R}} 

\newcommand{\vect}[1]{\mathbf{#1}} 

\IEEEoverridecommandlockouts
\newtheorem{remark*}{Remark}

\title{ContactNet: Online Multi-Contact Planning for Acyclic Legged Robot Locomotion}

\author{Angelo Bratta$^{1,2}$,
Avadesh Meduri$^2$,
Michele Focchi$^{1,3}$,
Ludovic Righetti$^2$,
and Claudio Semini$^1$
\thanks{$^1$ Dynamic Legged Systems (DLS) lab, Istituto Italiano di Tecnologia (IIT), Genova, Italy. Email: {\tt\small \href{mailto:name.surname@iit.it}{name.surname@iit.it}}}
\thanks{$^2$ Tandon School of Engineering, New York University (NYU), USA. Email: {\tt\small \href{mailto:am9789@nyu.edu}{am9789@nyu.edu}}, 
{\tt\small \href{mailto:ludovic.righetti@nyu.edu}{ludovic.righetti@nyu.edu}}.}
\thanks{$^3$ Department of Information Engineering and Computer Science (DISI), Università di Trento, Trento, Italy.}
\thanks{This work was supported by the European Union - NextGenerationEU, the Ministry of University and Research (MUR), National Recovery and Resilience Plan (NRRP), Mission 4, Component 2, Investment 1.5, project “RAISE - Robotics and AI for Socio-economic Empowerment” (ECS00000035), the European Union FSE-REACT-EU, PON Research and Innovation 2014-2020 DM1062 / 2021, and the National Science Foundation (grants 1932187, 1925079 and 2026479). }}
\begin{document}
	\maketitle
	\thispagestyle{empty}
	\pagestyle{empty}
\begin{abstract}
The field of legged robots has seen tremendous progress in the last few years. Locomotion trajectories are commonly generated by optimization algorithms in a \gls{mpc} loop. To achieve online trajectory optimization, the locomotion community generally makes use of heuristic-based contact planners due to their low computation times and high replanning frequencies.
In this work, we propose \textit{ContactNet}, a fast acyclic contact planner based on a 
multi-output regression neural network.
ContactNet ranks discretized stepping locations, allowing
to quickly choose the \textit{best feasible} solution, even in  complex environments.
The low computation time, in the order of 1 ms, enables 
the execution of the contact planner concurrently with a trajectory optimizer in a
\gls{mpc} fashion. 
In addition, the computational time does not scale up with the configuration of the terrain.
We demonstrate the effectiveness of the approach in simulation in different scenarios with the quadruped robot Solo12. 
To the best knowledge of the
authors, this is the first time a contact planner is presented that does not exhibit an increasing computational time on irregular terrains with an increasing number of gaps.
\end{abstract}
\section{Introduction}

Online motion planning for legged robots remains a challenging problem. The common approach is to use optimization algorithms in a Model Predictive Control (MPC) loop to automatically generate trajectories based on sensor feedback~\cite{DiCarlo2018, 
openaccess, Mastalli2023}. High frequency updates enable robots to react quickly to changes in the environment and reject external disturbances~\cite{Meduri2022}. In order to maximize the replanning frequency, the problem is often split into two components - contact planning and trajectory generation. Contact planning selects feasible footholds on the terrain to allow the robot to reach a desired location. Trajectory generation computes whole-body movements and contact forces to be applied at these locations.  
 
 Significant progress has been made in the area of online trajectory generation. Some approaches simplify the robot dynamics to a single rigid body with limited base rotations to render the underlying optimization problem convex~\cite{DiCarlo2018}. This allows for fast trajectory planning using a \gls{qp}. 
 Other methods use either the non-linear \gls{srbd} ~\cite{openaccess} or the full body model~\cite{Mastalli2023, Meduri2022} 
to generate almost any desired behavior on different robots.
Despite this progress, most of these methods still rely on heuristic-based contact planners~\cite{raibert1986legged} to ensure real-time computation. However,  such contact planners limit the overall motion planning framework (contact planning plus trajectory generation) to cyclic gaits only. 

The most recent works in this direction included the contact dynamics into a contact implicit MPC~\cite{kim2023contactimplicit, cleach2023fast}.
Kim et al.~\cite{kim2023contactimplicit} presented a DDP-based contact implicit \gls{mpc} with an analytical gradient for contact impulses to discover new gait sequences. However, no constraint on the feasibility of the chosen foot placement with respect to the morphology of terrain is considered.
A bi-level planning formulation was introduced in~\cite{cleach2023fast}.
The approach computes offline the optimal references and performs the linearization of the model for the entire motion; this allows to obtain an online local tracking controller that can change the contact plan in the presence of large disturbances.
Finally, \cite{Kong23} simulates in parallel multiple robot dynamics during the linesearch in the forward pass of a Hybrid LQR.
This allows to solve the hybrid system till convergence and thus obtain the best contact sequence.
The approach is computationally heavy and so it depends on the performance of the Isaac Gym simulator~\cite{isaacgym}.
However, automatically navigating terrain with constraints such as stepping stones is often not possible with such approaches. 
When complex motions are desired, the user is then forced to design a contact plan suitable for the desired task~\cite{Grandia23}.
 
\begin{figure}[!t] 
	\centering
	\includegraphics[scale=0.14]{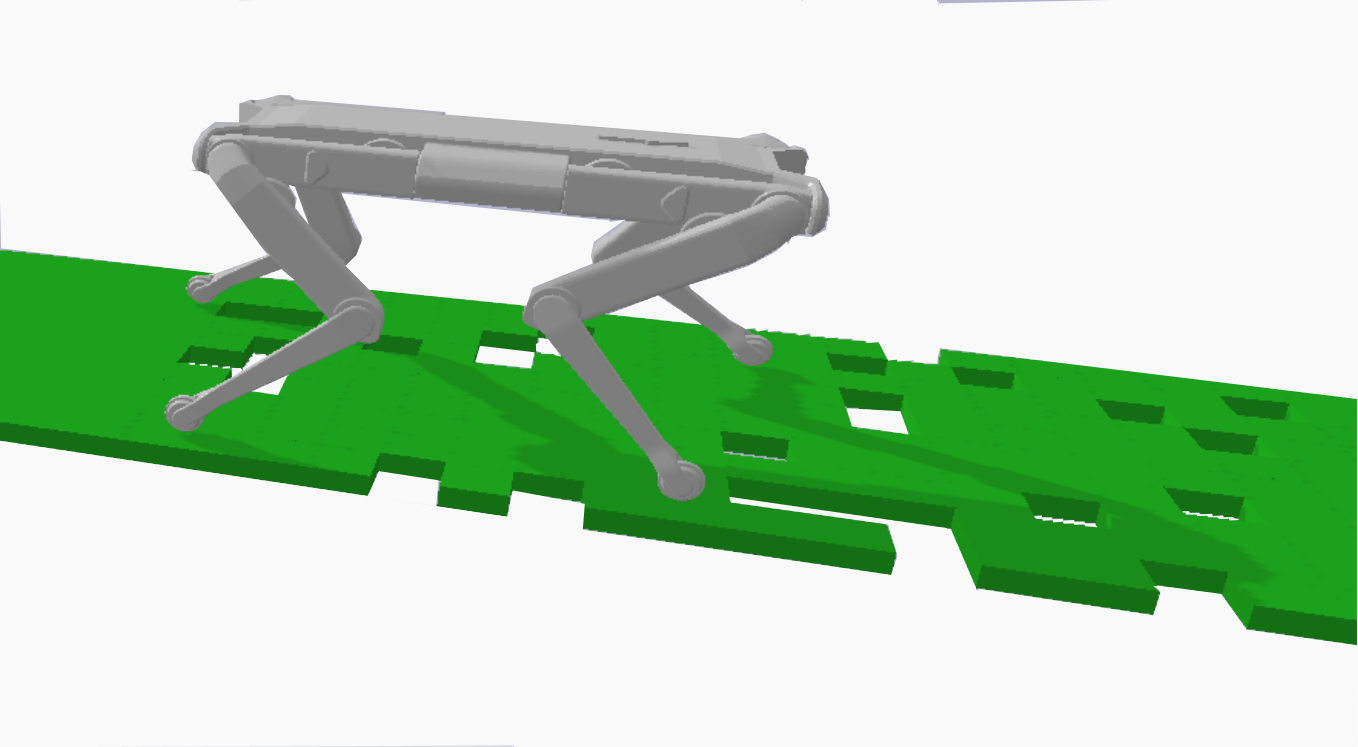}
	\caption{Solo12 robot traversing generated terrain with randomly removed squares of 5x5 $\mathrm{cm}$ dimensions.}
	\label{fig:solo}
\end{figure}

In the literature, there are examples of contact planners that can tackle complicated situations. Deits and Tedrake~\cite{Deits14} proposed a \gls{mip} to find footholds that avoid obstacles and violations of the kinematic limits. Similarly, the contact planning problem can also be optimized by maximizing the sparsity of the contact selection vector~\cite{tonneau2020sl1m}. Alternative to optimization techniques, sampling-based methods have been proposed to select feasible contact plans. 
For example, Lin et al.~\cite{Lin2020} presented a search-based footstep planner which explicitly takes into account disturbances. A neural network predicts if a  a candidate foothold location is zero-step
and one-step capturable for a full-body dynamic model.
Amatucci et al. \cite{Amatucci2022} presented a contact planner based on a Monte Carlo Tree Search (MCTS) algorithm; even though this approach 
demonstrated good performance, the expansion of the MCTS becomes too slow when a high number of discrete options are available,
e.g. terrain with holes. Consequently, all these methods cannot be used in an \gls{mpc} fashion at high control rates.

In this work, we address these limitations and propose an online, \gls{mpc} friendly multi-contact planner - \textit{ContactNet}, that can automatically generate arbitrary gait schedules, select footholds in unstructured environments, e.g. stepping stones, and recover from external perturbations. This contact planner extends the principles used in~\cite{meduri2021deepq}, a reactive planner for bipedal locomotion, which was limited to single contacts and cyclic gaits. ContactNet, on the other hand, computes acyclic gaits online for multiple legs. 
The key point of this approach is that the solve time is low and remains unaffected by the number of terrain constraints, such as stepping stones. 
  
ContactNet is based on a multi-output regression network \cite{Schmid22} that ranks a discrete set of foothold locations. This information is then used to generate a contact plan. The ContactNet is trained offline on a simple flat terrain using data generated with a novel cost function (see Sec. \ref{sec:costFuction}) which considers robustness, stability and minimizes trajectory generation cost. After training, we combine the ContactNet foothold plan with a centroidal trajectory optimizer~\cite{openaccess} to generate online a desired behavior. 

To evaluate our approach, we generate \textit{acyclic walk} and 
\textit{acyclic trot} behaviors on the Solo12 robot~\cite{Grimminger2020} in simulation (Fig. \ref{fig:solo}). We show that ContactNet can automatically navigate terrains with holes, even though those kind of terrains are not considered during the data collection for training. 
Finally, we systematically analyze the robustness of the ContactNet in face of measurement uncertainties, i.e. Gaussian noise in the joint velocity measurements, to emulate the behaviour of a real sensor.

\subsection{Contributions}
In summary, this paper proposes a fast contact planner for legged locomotion with the following main contributions:
\begin{itemize}
    \item the \textit{ContactNet}, a neural network-based contact planner, which can rapidly generate acyclic gait sequences
with safe footholds, even in the presence of holes in
the ground, considering tracking performance of an \gls{mpc}, stability
and robustness. To the best knowledge of the
authors, in contrast to all the other state-of-the-art approaches that suffer from terrain complexity,
our contact planner is the only one with a computational time that does not increase with the number of gaps present in the terrain.
    \item extensive preliminary simulation results with Solo12 that demonstrate the effectiveness of our approach with two gaits: an \textit{acyclic walk} and an \textit{acyclic trot} to navigate terrains with constraints (stepping stones). 
    We show that changing online the gait sequence is  crucial to address certain situations where fixed gait sequences fails.
\end{itemize}

\subsection{Outline}
The paper is organized as follows: Sec. II presents other works related to the proposed approach. Sec. \ref{sec:contactPlanner} gives an
overview of our contact planner.
Sec. \ref{sec:results} presents the results of simulation with the Solo12 robot with ContactNet for both walk and trot in different scenarios.
Finally, limitations and conclusions are drawn respectively in Sec. V and VI.

\section{Related Work}
Several motion planning methods that handle both contact planning and trajectory generation together by solving a non-linear problem have been developed in the past years. Posa et al. \cite{posa2014direct} use complementary constraints to ensure that that the end-effector either moves or applies a force to the environment. Winkler et al.~\cite{Winkler2018a} presented a trajectory optimization formulation which considers also foot position and stance/swing duration to generate different gaits. 
Ponton et al. \cite{ponton2021efficient} use Mixed-Integer Quadratically Constrained QPs to find contact sequences and whole-body movements for humanoids. On the other hand, methods exist that use convex optimization. Aceituno-Cabezas et al. \cite{Aceituno18} use Mixed-Integer Convex Programming to plan for both \gls{com} trajectory and contacts for the quadruped robot HyQ~\cite{semini11hyqdesignjsce}.
Recently, Jiang et al.~\cite{Jiang23} obtained a QP formulation to compute optimal trajectories of the \gls{com} by neglecting several terms in the centroidal dynamics. In addition, the authors extended the approach to an \textit{offline} Mixed-Integer QP which plans also for 
gait sequences, timings, and foot locations.  
A common drawback of all these methods is that they are not fast enough to be used in an \gls{mpc} fashion, which is important to compensate for model inaccuracies and external disturbances.

As already mentioned, in order to reduce the computational effort, several approaches assume a predefined gait sequence and optimize only the foot locations.
For example, Villarreal et al. developed a foothold classifier based
on a Convolutional Neural Network (CNN)~\cite{villarreal19ral} and combined it with a \gls{mpc}-based trunk controller \cite{Villarreal2020}
to achieve reactive and real-time obstacle negotiation, considering a 3D map of the terrain.
Another example of using a CNN to select the optimal landing location was presented by Belter et al.~\cite{Belter19}. 
Their approach is based on the learning of a model to evaluate the quality a potential touchdown point taking into account the local elevation map, kinematic constraints and collision.
Grandia et al.~\cite{Grandia23} 
performed a convex inner approximation of the steppable terrain, optimizing foot locations inside that region.
However, the authors mention that a change in the gait sequence
should be required to prevent the robot from falling in the presence of strong disturbances.

In Section IV we showcase a scenario in which computing online both footholds and gait sequence is fundamental to accomplish the motion.

\section{ContactNet} \label{sec:contactPlanner}

ContactNet computes foot locations and contact status (i.e., swing or stance) for each leg in the horizon.
In this section, we describe the cost function and data generation approach used to rank footholds offline. After that, we discuss the details regarding ContactNet and we present the entire framework used to generate acyclic multi-contact plans. 

\begin{figure*}[!t] 
	\centering
	\includegraphics[scale=0.39]{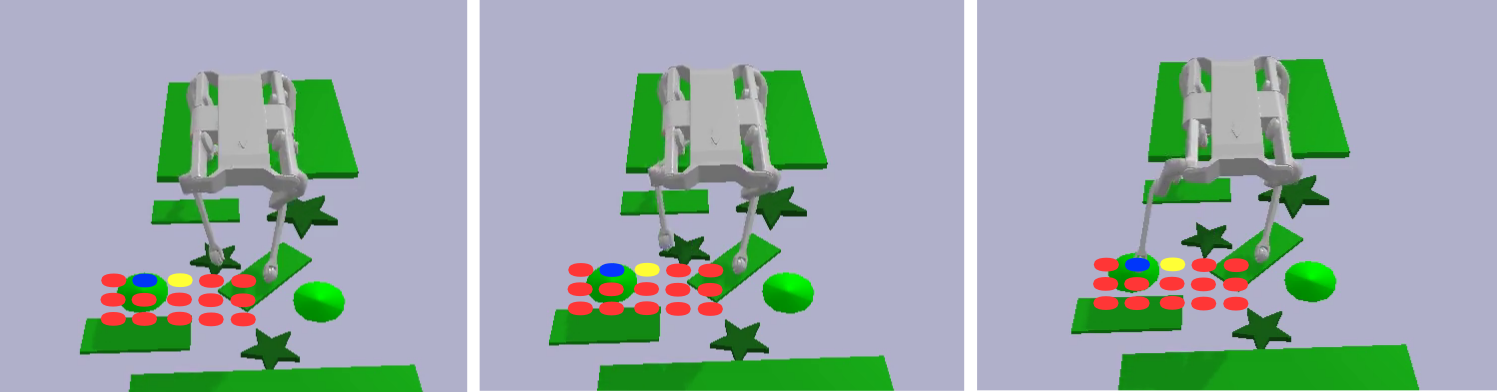}
	\caption{Example of the evaluation on the ContactNet on a terrain composed of stepping stones. 
	Red disks represent some of the actions evaluated by the ContactNet. The others are not shown for image clarity.
 The network computes the ranking order, according to which the yellow disk is the one which minimizes the cost function \eqref{eq:cost}. In this example, knowing the terrain map,  yellow disk is discarded because it corresponds to an hole in the terrain. 
 Checking iteratively in the ordered output of the ContactNet, the blue disk is chosen since it corresponds to the first action deemed \textit{safe}.}
	\label{fig:snapshot}
\end{figure*}

\subsection{Footholds}
We discretize the allowed stepping region for each leg into a fixed set of $N_\mathrm{a}$ possible locations. These footholds are defined at fixed distances from the current hip location of the corresponding foot, similar to~\cite{meduri2021deepq}.
Subsequently, as the robot moves, the allowed foot locations also change. 
Discretizing the candidate footholds is quite a common approach, e.g., \cite{villarreal19ral}; we show in our experiments that, despite losing the freedom of stepping anywhere in the feasible region very reliable behaviors can be generated.
 
\subsection{Cost Function} \label{sec:costFuction}
Given the discrete set of possible footholds for all the legs, the goal is to identify the best one, considering the morphology of the terrain, the references and the current state of the robot. 
For this, we propose a novel cost function that is used to rank all the foot locations based on several aspects, such as robot stability, robustness and optimal trajectory. We consider the input to be:
\begin{equation}
\vect{u}_\mathrm{r} = [{}_\mathcal{C} \vect{p}_\mathrm{f}, \vect{p}_\mathrm{c, z}, \vect{v}_\mathrm{c}, 
\vect{v}_\mathrm{c}^\mathrm{usr} ] \label{eq:input}   
\end{equation}
where  
${}_\mathcal{C} \vect{p}_\mathrm{f} \in \Rnum^8$
represents X and Y components of the foot location in the \gls{com} frame $\mathcal{C}$\footnote{All the quantities without left subscript 
are expressed in the inertial fixed World frame $\mathcal{W}$.},
$\vect{p}_\mathrm{c,z} \in \Rnum$ is the Z component of the \gls{com} position,
$\vect{v}_\mathrm{c} 
\in \Rnum^3$ is the actual \gls{com} velocity.
Finally, the variable $\vect{v}_\mathrm{c}^\mathrm{usr} 
\in \Rnum^2$ is the user-defined reference linear velocity. 

To evaluate a foothold, we first generate a trajectory that moves the robot from the current configuration to the chosen one and then use the following cost function
\begin{equation} \label{eq:cost}
V = 
\sum_{k=0}^{N_\mathrm{s}} V_\mathrm{k} + V_{N_\mathrm{s}}
\end{equation}
where $N_\mathrm{s}$ is the \textit{step horizon}, $V_\mathrm{k}$ is the running cost (evaluated at each node of the trajectory), $V_{N_\mathrm{s}}$ is the terminal cost (evaluated only at the final point). 
The running cost $V_\mathrm{k}$ consists of three terms
\begin{equation}
V_\mathrm{k} = \gamma_{opt} V_\mathrm{k,opt} + 
\gamma_{stab} V_\mathrm{k, stab} + \gamma_{kin} V_\mathrm{k, kin}.
\end{equation}
The first term corresponds to the cost of the optimization problem obtained from the trajectory optimization~\cite{openaccess},
i.e. tracking of references for states 
(\gls{com} quantities)
and control inputs (\gls{grfs}).
It guarantees that a feasible trajectory that respects the dynamics and 
friction cone constraints exists. 
In this work, we
use a \gls{srbd} model \cite{Orin2013a}, but any other model could also be used.
The variable $V_\mathrm{k, stab}$, evaluates the margin of stability of the motion. It computes the distance of the projection of the \gls{com} on the ground from the closest support polygon edge. 
For instance, in a walk, this encourages footholds in which 
the robot is statically stable (\gls{com} inside the support polygon, $V_\mathrm{k, dist}$ = 0);
for a trot, this maximizes the controllability of the robot.
The last term $V_\mathrm{k, kin}$ enforces kinematic limits - it assigns a high value when a leg in stance violates these limits. Even though our simplified model does not include joint values, we consider a violation of the kinematic limits if the distance between the foot and the hip exceeds a certain threshold. 
To do so, we assume that the positional offset between the hip and the \gls{com} remains constant for the entire trajectory. 
Further, a conservative threshold value is chosen to encourage the motion of one leg when it is close to the kinematic limits, i.e. to place it in a more kinematically favourable location, similarly to \cite{Bjelonic21}.

The terminal cost $V_{N_\mathrm{s}}$ in the cost function $V$ takes into account future actions of the robot. It is defined as follows:
\begin{equation}
V_{N_\mathrm{s}} = \gamma_{cent} V_\mathrm{cent} - \gamma_{area} V_\mathrm{area}.
\end{equation}
With $V_\mathrm{cent}$, we introduce a penalization on the distance between the projection of the \gls{com} and the center of the support polygon.
Minimizing this quantity increases the number of subsequent stable steps.
Finally, the quantity $- V_\mathrm{area}$ improves the robustness of the contact configuration by maximizing the area of the final support polygon. 

The numbers $\gamma_i \in \Rnum$ scale the different cost terms.

\subsection{ContactNet}
\begin{figure*}[!t]
	\centering
	\includegraphics[scale=0.64]{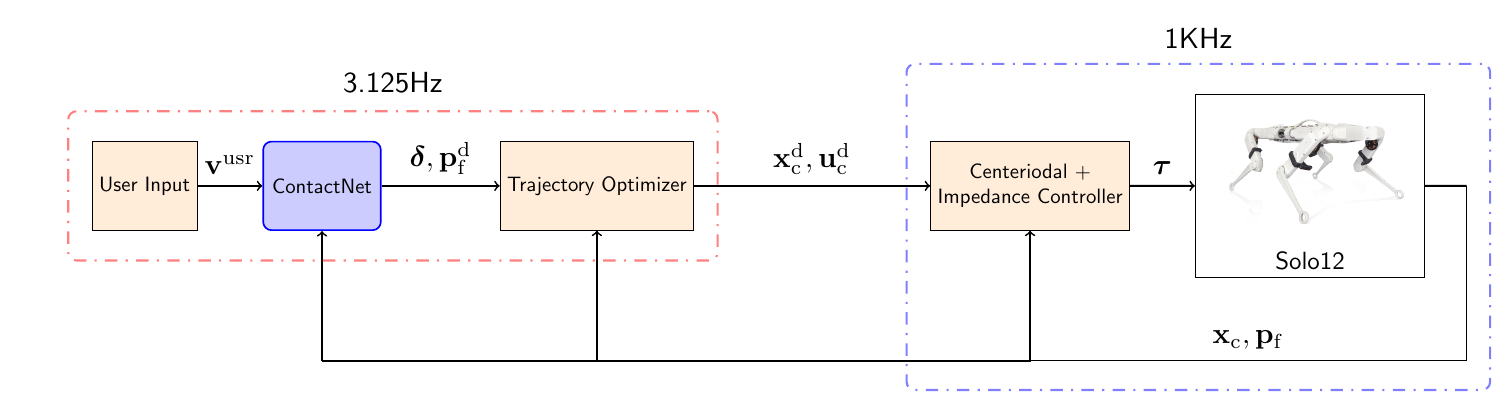}
 \caption{Block scheme of the entire locomotion framework. Given the user-defined velocities $\vect{v}^\mathrm{usr}$ the actual robot state 
 $\vect{x}_\mathrm{c}$, and actual foot locations $\vect{p}_\mathrm{f}$ the ContactNet computes in a few milliseconds the gait sequence $\boldsymbol{\delta}$ and touchdown points $\vect{p}_\mathrm{f}^\mathrm{d}$ for the three following steps, at a frequency of 3.125 Hz (after each touchdown). Given the sequence as parameter, the Trajectory Optimizer~\cite{openaccess} computes \gls{com} trajectory and \gls{grfs} tracked by a 1 $\mathrm{kHz}$ centroidal whole-body controller and a joint space impedance controller \cite{Grimminger2020}.
In order to guarantee that the motions are feasible also on the real robot, the torques are saturated to the maximum values that the motor of Solo12 can produce.}  
	\label{fig:scheme}
\end{figure*} 
Using the cost function discussed previously, it is possible to automatically generate acyclic multi-contact plans for locomotion by simply selecting as \textit{action} - which leg to move and where to step - the candidate with lowest value of $V$. 
However, evaluating all the possible footholds by computing optimized trajectories is not feasible online.
Consequently, we propose to train offline a neural network that learns to rank the possible footholds using the cost function \eqref{eq:cost}, giving the input of~\eqref{eq:input}.

\subsubsection{Data Generation}
To train the ContactNet, we generate a dataset containing many possible stepping situations that the robot can be dealing with on a flat terrain. We start the robot in a randomly generated configuration (different joint position and velocity) and choose a random reference \gls{com} velocity in the range (\mbox{-0.1},0.1 $\mathrm{m/s}$) for both X and Y directions. 
Before each liftoff, the cost function \eqref{eq:cost} is evaluated, and the best foothold among the discrete options is selected, i.e. the one with the smallest $V$, is selected. 
Subsequently, a trajectory is generated with this contact plan and is tracked on the robot in simulation. We define this as an \textit{instance}. After that, a new instance is run (same reference velocity, starting from the configuration achieved at the end of the previous instance) to generate a large dataset containing the input $\vect{u}_\mathrm{r}$ and the corresponding $\vect{V}$, i.e., the vector which contains the values of cost $V$ for each option. 
A new \textit{episode} is restarted (i.e., new reference velocity and initial configuration) after 30 instances or when the robot falls down. 
In this way, the dataset contains the configurations the robot will likely have during a motion.
Since we do not know which step led to the final fall, we heuristically remove the last 3 instances of the episode in case of falling.
The framework is not real-time friendly, but generating the dataset does not take too long - about 6-7 hours with a standard computer.

\subsubsection{ContactNet Training}
Our goal now is to learn a function $f_\theta: 
\Rnum^{14} \rightarrow \Rnum^{N_\mathrm{a}}$, which maps the current input  $\vect{u}_\mathrm{r}$ to the list of the ranked footholds. 
The main advantage of a learning approach is that it guarantees low computational effort at runtime, allowing us to integrate it with our \gls{mpc}.

In order to learn the ranking function for all the possible actions, we used a multi-output regression network \cite{Schmid22}.
To do so, we sort the vector $\vect{V}$ from the dataset in decreasing order 
and create a vector $\vect{Y} \in \Rnum^{N_a}$, assigning to each value of $\vect{V}$
its index in the sorted vector;
the smaller the cost for the action $a$, 
the higher its value in $\vect{Y}$. 
We normalize each entry by $N_\mathrm{a}-1$, such that the values in $\vect{Y}$ are between 1 and 0 (1 smaller cost, 0 higher cost).
For the sake of clarity we provide a small example, i.e., an instance with only 3 possible footholds:
$\vect{V} = [0.8, 0.3, 0.9]$, 
sorted($\vect{V}) = [0.9, 0.8, 0.3]$ 
$\vect{Y} = [1/2, 2/2, 0/2]$, since the cost 0.8 has index 1, the cost 0.3 has index 2, and the 0.9 has index 0 in the sorted $\vect{V}$.

As a training loss, we use the mean squared error between the prediction $\vect{\hat{Y}}$ = 
$f_\theta(\vect{u}_\mathrm{r})$ and $\vect{Y}$.
\begin{equation}
\displaystyle{\min_{\theta}}\quad \frac{1}{N_\mathrm{a}}
\sum_{i=0}^{N_\mathrm{a}-1}  ( \vect{Y}_\mathrm{i} - 
f_\theta(\vect{u}_\mathrm{r})_i )
\end{equation}
\subsubsection{ContactNet Evaluation}
After training the ContactNet, we can quickly obtain the optimal foothold by constructing the vector of indices that sort in decreasing order $\vect{\hat{Y}}$ (we refer to this vector as $\vect{\hat{Y}'}$) and pick the first value. 
In the example before, assuming a perfect output of the neural network, we have
$\vect{\hat{Y}} = [1/2, 2/2, 0/2]$, and so
$\vect{\hat{Y}'} = [1, 0, 2]$. This means that the action number 1 in the discretized set is the optimal one. 
Even though we consider only flat terrains, in certain situations some options in $\vect{\hat{Y}'}$ must be discarded because \textit{unsafe}, i.e. they correspond to a point in which there is a hole in the terrain. 
We identify if a foothold is safe by using the knowledge of the terrain map.
In particular, we iteratively check the elements in the vector $\vect{\hat{Y}'}$ till we find the one that does not coincide with a hole. 
Since $\vect{\hat{Y}'}$ has been ordered based on the lower cost function, this approach chooses the optimal safe action (which leg to move and where to place the foot).
An example of such a situation is shown in Fig. \ref{fig:snapshot}, where the robot is expected to walk across stepping stones. The red circles correspond to some of the allowed footholds of the Left Front leg. The others are not shown for image clarity but are also evaluated in this situation.
The yellow disk represents the first value in $\vect{\hat{Y}'}$, but it cannot be selected since there is no terrain below it. 
Consequently, we check the following elements in $\vect{\hat{Y}'}$ till the first one that is coherent with the terrain, e.g. blue disk.

\textit{Remark:}
We chose to discretize the foothold locations and rank all of them, mainly to navigate complicated terrain situations online without adding the morphology of the terrain directly into the formulation. 
\subsection{Overall control architecture}
\label{subsec:control_architecture}
\begin{figure*}[!t]
	\centering
	\includegraphics[scale=0.39]{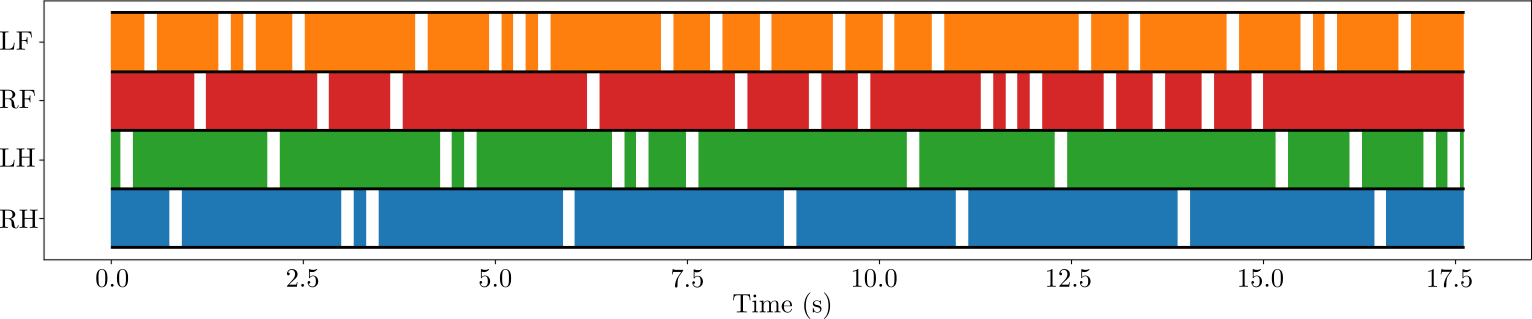}
	\caption{Gait schedule of a walk motion on a stepping stones scenario. White parts indicates moment in which that leg is in swing. The ContactNet finds a completely acyclic gait.}
	\label{fig:steppingStones}
\end{figure*}
Figure \ref{fig:scheme} shows the block scheme of our locomotion framework. 
The user decides the linear velocities $\vect{v}_\mathrm{c}^\mathrm{usr} 
\in \Rnum^2$ that the robot should follow. 
Given the X and Y components of actual foot position in the \gls{com} frame 
$\mathcal{C}$ ${}_\mathcal{C} \vect{p}_\mathrm{f} \in \Rnum^8$, 
Z component of the actual \gls{com} position $\vect{p}_\mathrm{c,z}$, 
actual \gls{com} velocity $\vect{v}_\mathrm{c}$ 
and reference velocities $\vect{v}_\mathrm{c}^\mathrm{usr}$, 
the ContactNet returns the best candidate foothold, as explained in the previous section.
In our architecture we compute online a contact plan with three steps for a prediction horizon of $N$ = 3 $N_\mathrm{s}$ step horizons.
Reference velocities are integrated to compute the \gls{com} position at the end of each step horizon.
They  are used together with the chosen foothold 
to define the input $\vect{u}_\mathrm{r}$ of \eqref{eq:input} for the second step horizon to re-evaluate the neural network; similarly it happens for the third evaluation.
The swing times are preset depending on the chosen gait (discussed in detail IV-A). This contact plan along with the reference \gls{com} trajectories and the reference \gls{grfs}\footnote{Weight of the robot divided by the number of legs in contact with the terrain for the stance phase, zero for the swing phase} are provided to the trajectory optimizer to generate an optimal movement using the algorithm described in~\cite{openaccess}. 
The \gls{com} trajectories are then tracked by a 1 $\mathrm{kHz}$ whole-body controller~\cite{Grimminger2020},
combined with a PD controller in Cartesian space for the swing trajectories. 
The swing trajectory is defined in the swing frame; 
a semi-ellipse represents the X component
and a fifth-order polynomial the Z.
At the end of each step horizon, the procedure is repeated in \gls{mpc} fashion.
\section{Results} \label{sec:results}
In this section, we present the results obtained by our approach. We perform simulations with Solo12, 
a 2.2 $\mathrm{kg}$ open-source torque-controlled modular quadruped robot.
The entire framework runs on a Dell precision 5820
tower machine with a 3.7 GHz Intel Xeon processor. We perform our simulation using the PyBullet library~\cite{coumans2019}.

For all the experiments, the ContactNet is composed of 4 fully connected layers with $128$ neurons each. All layers except the last one are
activated with a ReLU function. 
As hyper-parameters for training, we choose a number of epochs equal to 1000
with a batch size of 100. The learning rate is set to 0.001. 
The input $\vect{u}_\mathrm{r}$ is normalized to be in the range (\mbox{-1},1) to improve the accuracy of the network~\cite{LeCun2012}.
To evaluate the network's performance we used 70 \% of the entries of the dataset as a training set and the remaining part as a test set. We use a top-5 metric to determine the statistics of the network, i.e., we consider a correct prediction if the first element of $\vect{\hat{Y}}$, i.e. what the neural network outputs as a best action, is one of the first five elements in the corresponding $\vect{V}$ stored in the dataset.
In our case, this metric has a particular importance since the best action will not be always feasible due to the requirements of the terrain. 
\subsection{Acyclic gaits}
In this subsection, we discuss the various parameters defined to generate the two gaits - walk and trot. 
\subsubsection{Walk}
In this experiment, the robot is only allowed to move one leg at a time.
We choose a discretization time of 40 $\mathrm{ms}$ for the trajectory optimization.
The step horizon $N_\mathrm{s}$ is equal to 320 $\mathrm{ms}$ (8 Nodes) and it is composed of 
120 $\mathrm{ms}$ of four leg stance phase (3 nodes), 
160 $\mathrm{ms}$ (4 nodes) of swing phase, and the last node of four leg stance phase.
The prediction horizon $N$ used by the trajectory optimizer 
is composed of three step horizons, 960 $\mathrm{ms}$, corresponding to three evaluations of the neural network, as discussed in Sec. \ref{subsec:control_architecture}.
The duration of the swing and stance phase has been chosen based on our previous experiments with the Solo12 robot; the presented approach is generic and can be applied with other values for swing/stance.

We define the allowed stepping region for each leg to be a 20 $\times$ 20 $\mathrm{cm}$ grid which is a meaningful size given the kinematic limits of Solo12. This space is discretized into 25 footholds which are 5 $\mathrm{cm}$ apart, see red disks in Fig.~\ref{fig:snapshot}. Subsequently, the network needs to choose among a total of $N_a$ = 100 possible footholds (4 × 25) since we do not prescribe which leg needs to swing, but we only require one leg swing at a time.
For data generation, we run 1500 episodes using the procedure discussed in Sec. \ref{sec:costFuction}. The resulting data had 43410 instances of ($\vect{u}_\mathrm{r}/\vect{V}$) tuples. 
We obtained an accuracy of the 93.48/90.81 \% in the training/test set according to the top-5 metric.

\subsubsection{Trot}
In the trot gait, two diagonal feet are leaving the ground at the same time.
The total stepping region for each leg is a square size 10 $\times$ 10 cm. The foothold discretization resolution is still 5 $\mathrm{cm}$, 9 choices per leg. At the start of a stepping horizon, there are a total of $162$ - $2 \times 9^2$ foothold choices since at each step two legs leave the ground. All the other parameters are the same as the walk. 
We run 1500 episodes to generate the dataset for this gait and train the ContactNet, obtaining 45000 instances. The neural network achieves an accuracy of 
99.48/97.7 \% in the training/test set.

In the accompanying video\footnote{\url{https://www.youtube.com/watch?v=ta1JpSigRKo}}, we show a long horizon trot motion in a scenario with holes in the terrain. The reference velocity changes every 10 s in the range $(-0.1, 0.1) \mathrm{m/s}$. In this way, we demonstrate the locomotion stability and the ability of avoiding unsafe footholds of ContactNet.

\subsection{Stepping stones scenario}
\label{subsec:stepping_stones}
To verify the effectiveness of our \gls{mpc} framework, we designed a terrain composed of 8 sparse stepping stones of different shapes: 
3 stars, 2 circles and 3 rectangles, see Fig. \ref{fig:snapshot}.
Two squares are positioned as starting and end points. The goal of the task is to traverse the terrain with a user-defined forward velocity of 0.05 $\mathrm{m/s}$ using the ContactNet trained for the walk gait. The proposed approach successfully navigates the terrain. Fig. \ref{fig:steppingStones} shows the resulting gait sequence for the entire motion. Each color corresponds to a different leg; white parts indicate that the foot is in the air at that time. 
Fig.~\ref{fig:steppingStones} demonstrates that the motion is completely acyclic. For example, when the front legs are on the last square the \gls{com} is closer to the hind legs, automatically making the swing of the front legs preferable to prevent the \gls{com} from going outside the support triangle. The hind legs are moved only when stability is guaranteed and the set of actions contains a touchdown point on a stepping stone. The result of the simulation is shown in the accompanying video. The ContactNet in average has chosen the 6th element of $\vect{\hat{Y}}$, with a maximum of 30 discarded elements
for the swing of the Left Hind leg at time around 10 $\mathrm{s}$.
The average computation time for a complete iteration of the ContactNet, i.e. computation of 3 subsequent actions/footholds, is
1.6 $\mathrm{ms}$, which is much faster than the trajectory optimizer. This scenario is particularly challenging for contact implicit MPC~\cite{kim2023contactimplicit} since the morphology of the terrain should be considered in the formulation. Similarly, a sampling-based approach as~\cite{Amatucci2022} would suffer from the exponential increase of computation time.

\begin{figure*}[!t]
 	\includegraphics[scale=0.51]{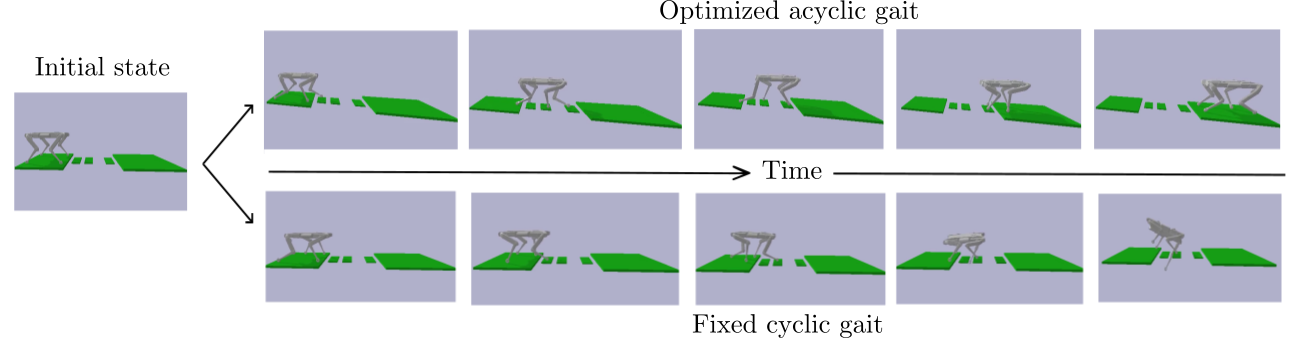}
	\caption{A comparison between fixed cyclic and optimized acyclic gait selection: Solo12 fails to traverse a narrow stepping stone scenario when ContactNet is only allowed to adapt foot locations. Solo12 traverses the terrain when adapting both footholds and gaits online.}
	\label{fig:steppingStones2}
\end{figure*}

To demonstrate that the ContactNet can be generalized to any stepping stones scenario we considered a second terrain with three rectangles of different sizes, see Fig.~\ref{fig:steppingStones2}. The terrain is designed to make the stepping stones narrow and spaced unevenly with the last stepping stone farthest from the previous. Crossing this terrain would require optimal foot location planning. Additionally, we use this terrain to evaluate the impact of online gait selection along with pure foot location adaptation.
We modified the evaluation procedure of the ContactNet in order to be able to find the corresponding optimal safe foothold for a specific leg. In such a case ContactNet only adapts foot location as it happens in approaches such as~\cite{villarreal19ral}, using a fixed gait sequence.
We initialize Solo12 with the same initial state and let it traverse the terrain twice, once with the original ContactNet able to choose both the footholds and the gait sequence (\textit{optimized acyclic gait}), and then using ContactNet with \textit{fixed cyclic gait}, 
In this case, when none of the discretized options for a leg are coherent with the terrain, a swing in place is forced.
Fig \ref{fig:steppingStones2} shows frames of the behaviour of Solo12 in the two scenarios. Solo12 fails to traverse the terrain when it is not allowed to adapt the gait online. In the case where Solo12 chooses gait and foothold it is able to navigate safely across the terrain by finding a feasible and stable contact plan. 
This is an example that demonstrates the need for both acyclic gait and foot location selection (as it is done by the ContactNet) to navigate complex terrain.    
\subsection{Randomly generated terrain}

\begin{figure}
	\centering
	\includegraphics[scale=0.17]{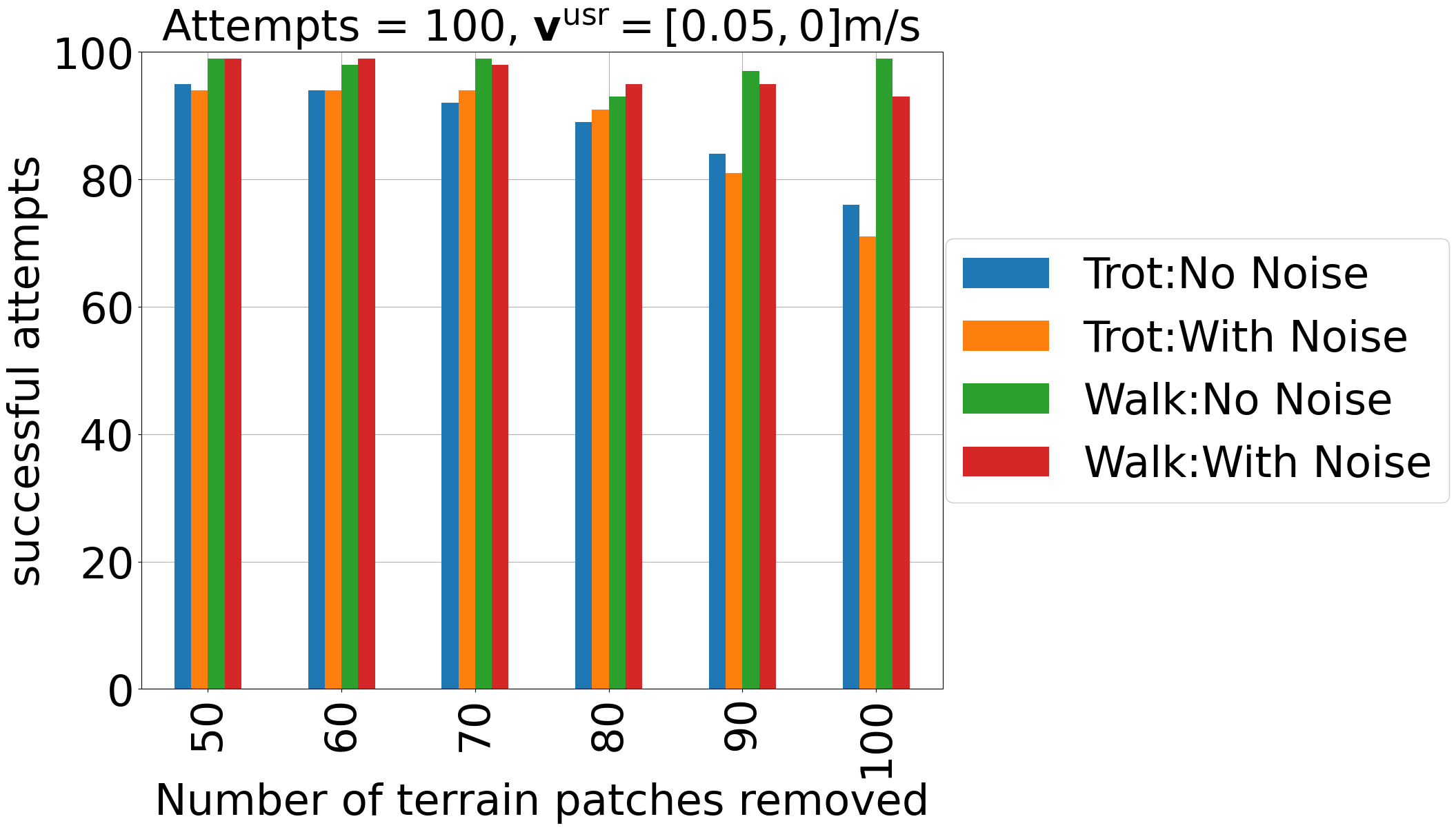}
	\caption{Successful attempts of both walk and trot motions over the number of 
         terrain patches removed with. For each number of terrain patches we execute 100 trajectories with and without noise (Gaussian noise with zero mean and 0.01 variance). }
	\label{fig:random}
\end{figure}

In this section, we evaluate the reliability of our \gls{mpc} scheme to navigate unstructured environments with various terrain constraints. We generate a terrain of 1.5x0.5 $\mathrm{m}$ by placing 300 squares 5x5 $\mathrm{cm}$. Starting from a number of $n$ = 50 till $n$ = 100 (17 \% - 33 \%),  we randomly remove $n$ terrain patches and evaluate the success rate for both the walk and trot setups (see Fig. \ref{fig:solo}) with a reference velocity of 0.05 $\mathrm{m/s}$ in the X direction. For each $n$ we performed 100 different attempts, changing the terrain configuration. A trial is considered successful if the robot reaches the last terrain patch. In addition we also repeat the navigation task for the two gaits with Gaussian noise (zero mean, 0.01 variance) applied to joint velocity measurement. This is performed to validate the robustness of the approach in conditions closer to the real robot. The results are shown in Fig. \ref{fig:random}.

The ContactNet has a high rate of success for both gaits, while, as expected, the walk guarantees better performance due to its intrinsic stability.
The addition of noise does not cause a significant reduction in performance. This suggests that our framework would reliably work on a real robot. Note that the ContactNet remains robust to noise even though it was not trained for it, as is commonly done using domain randomizing techniques~\cite{Tobin17}. Further, no assumptions are made on how terrain patches are removed to guarantee that a real feasible path exists. 

Another important element of our analysis is the solve time of the ContactNet. We consider the total solve time to be 3 evaluations of the network, i.e., the time needed to generate the contact plan for our trajectory optimization. 
Table \ref{tab:comp_time} reports the mean value of the computational time of ContactNet for each value of $n$ during one attempt.
We highlight that the time is low, around 1 $\mathrm{ms}$, and does not change with the complexity of the terrain. This contrasts with other approaches, such as \gls{mip} and MCTS, where the computational time suffers from the dimension of the solution space.
For example, the MIP of \cite{Jiang23} takes 55789.3 ms an average for a full walking cycle thus requiring to be solved offline; the MCTS of~\cite{Amatucci2022} requires an average of 400 ms to obtain a sequence of 6 steps, given the high number of possibilities and it is already 2.72 faster than a MICP similar to the one presented~\cite{Aceituno18}.
In addition, the non-convexity of the star stones is hard to be considered in an analytic constraint.
Furthermore, ContactNet takes into account the morphology of the terrain, differently from the online contact implicit MPCs such as~\cite{kim2023contactimplicit}.
The ContactNet, indeed, only has to query the terrain map until the first safe action is found.
 
\begin{table}[h!]
    \caption{Average Computational time of ContactNet in one attempt in Randomly Generated Terrain}
	\begin{center}
		\begin{tabular}{@{} l l l l @{}}
			\toprule[0.4mm]
			\textbf{Number of terrain patches removed} & {\textbf{Walk [ms]}} & \textbf{Trot [ms]} \\ 
			\midrule
			50 & 1.272&  1.507     \\
			60 & 0.938&  7.950    \\
			70 & 0.944&  0.803   \\
			80 & 1.152&  0.785      \\
			90 &1.0584&  0.874     \\
			100 &0.9167&  0.7899      \\
			\bottomrule[0.4mm]
		\end{tabular}
		\label{tab:comp_time}
	\end{center}
\end{table}
As done for the stepping stones scenario (Sec.~\ref{subsec:stepping_stones}) we analyzed the difference between changing online the gait sequence and using a fixed gait during the walk. Table~\ref{tab:fixed_vs_acyclic} reports the result of 300 attempts when $n=110$ blocks have been removed. To achieve a fair comparison, the scenario was created randomly, but both approaches were run on the exact same test scenarios. The success rate for ContactNet with \textit{optimized acyclic gait} is 93.6\% (only 19 attempts failed), while when using a \textit{fixed cyclic gait} the robot is able to accomplish the task only in 85.3\% of the cases (44 fails, more than twice than the previous case). 
This result confirms that an online acyclic gait planner has better performance than a foothold adaptation algorithm with a fixed gait.
\begin{table}[h!]
	\caption{Comparison of success rate of fixed and acyclic gate in Randomly Generated Terrain with n=110 patches removed}
	\begin{center}
		\begin{tabular}{@{} l l l l l @{}}
			\toprule[0.4mm]
		\textbf{Type of gait} & \textbf{Attempts} & \textbf{Successful attempts}  & \textbf{Success rate}\\ 
			\midrule
			\textit{optimized acyclic} & 300 &  281 & 93.6 \%    \\
   			\textit{fixed cyclic} & 300 &  256 & 85.3 \%    \\
			\bottomrule[0.4mm]
		\end{tabular}
		\label{tab:fixed_vs_acyclic}
	\end{center}
\end{table}
\subsection{Push Recovery}
As a final result, we tested the robustness of the ContactNet pushing the robot for 1 $\mathrm{s}$ with external forces in the range of $\pm$ 5 $\mathrm{N}$ (25\% of the weight of Solo12) in both directions while tracking a forward velocity. In addition some terrain patches of 10x10 $\mathrm{cm}$ are randomly removed. While being pushed, the Contact Planner adjusts footholds to counteract the external disturbance and avoid holes. Once the push is removed, the robot automatically recovers a stable configuration, for example by first moving a leg which resulted to being close to the kinematic limits.

\section{Limitations}
In this paper we have presented preliminary results in simulation of an online multi-contact planning framework that can be easily integrated with existing trajectory optimization approaches.
Our analysis of terrain and sensor noise shows that the results have the potential to be transferred to the real robot. 
Even though the cost function~\eqref{eq:cost} does not consider torque limits, the joint space impedance controller guarantees that torques sent to the robot satisfy the torque limits by saturating them. 
Furthermore, the current formulation has been shown only on flat terrain. The ContactNet can be extended to uneven terrain by discretizing the 3D stepping region and retraining the network. While the trajectory generator could handle uneven terrains~\cite{openaccess}, we did not pursue this direction because uneven terrain locomotion requires additional components, such as a collision-free swing trajectory, which was not readily available and goes beyond the scope of the contact planning problem. 
Currently, the ContactNet does not update the foothold during the swing phase. Throughout our experiments, we found this replanning frequency to be sufficiently robust to uncertainties in the environment. However, if the need arises for faster updates, the trajectory optimizer has been demonstrated to be able to compute the optimal trajectories with arbitrary initial contact configurations. The same cost function \eqref{eq:cost} can be used to rank footholds during the swing phase as well. 
Finally, ContactNet can choose small steps in a row with the same leg, especially for the trot. An energy-based cost term could be added to the cost function~\eqref{eq:cost} to prevent "useless" swings and facilitate natural motions.

\section{Conclusion}
In conclusion, we proposed a multi-contact planner, \textit{ContactNet}, capable of generating acyclic gaits, i.e., without a predefined leg sequence, in a few milliseconds, even in the presence of unstructured terrains.
Simulations with Solo12 robot are performed with walk and trot motion. Robustness is demonstrated by inferring measurement noise and applying external disturbance.
We demonstrated that an acyclic gait planner performs better than a planner that chooses only the foot locations with a fixed gait.
Future work will consider the transfer of the approach to the real hardware.

\bibliographystyle{IEEEtran}
\bibliography{root.bib}
\end{document}